\documentclass[10pt]{article}

\usepackage{graphics}
\usepackage{graphicx}

\usepackage[a4paper, left=35mm,right=35mm,top=34mm,bottom=34mm]{geometry}
\usepackage[utf8]{inputenc}
\usepackage[T1]{fontenc}
\usepackage[english]{babel}

\usepackage{enumerate}
\usepackage{graphicx}
\usepackage{hyperref}
\hypersetup{
    colorlinks=true,
    linkcolor=blue,
    filecolor=magenta,      
    urlcolor=cyan,
}
\usepackage{listings}
\usepackage{color}

\definecolor{dkgreen}{rgb}{0,0.6,0}
\definecolor{gray}{rgb}{0.5,0.5,0.5}
\definecolor{mauve}{rgb}{0.58,0,0.82}

\usepackage{mathtools,amsthm,amssymb,amsfonts}
\usepackage{algorithm}
\usepackage{algorithmic}
\makeatother
\theoremstyle{plain}

\theoremstyle{definition}

\theoremstyle{remark}

\usepackage{caption} 
\usepackage{fancyhdr}
\author{
  {\normalsize Marien Chenaud}\thanks{Correspondance: marien.chenaud@centralesupelec.fr} \footnotemark[3] \footnotemark[4]
  \and
  {\normalsize Fr\'ed\'eric Magoul\`es} \footnote{Correspondance: frederic.magoules@hotmail.com} \thanks{CentraleSup\'elec, Universit\'e Paris-Saclay, France.}
  \and
  {\normalsize José Alves} \thanks{Transvalor S.A. E-Golf Park, 950 Avenue Roumanille, 06410 Biot, France.}
}

\title{Physics-Informed Graph Convolutional Networks: Towards a generalized framework for complex geometries}
\date{}

\begin{document}
\maketitle
\thispagestyle{fancy}

\begin{abstract}
	\noindent Since the seminal work of \cite{raissi2019physics} and their \textit{Physics-Informed neural networks} (PINNs),
	many efforts have been conducted towards solving partial differential equations (PDEs) with Deep Learning models.
	However, some challenges remain, for instance the extension of such models to complex three-dimensional geometries,
	and a study on how such approaches could be combined to classical numerical solvers.
	In this work, we justify the use of graph neural networks for
	these problems, based on the similarity between these architectures and the meshes used in traditional numerical techniques for solving partial differential equations. After proving an issue with the Physics-Informed framework for complex geometries, during the computation of PDE residuals, an alternative procedure is proposed, by combining classical numerical solvers and the Physics-Informed framework. Finally, we propose an implementation of this approach, that we test on a three-dimensional problem on an irregular geometry.
\end{abstract}


\section{Introduction}
\label{sec:intro}

Partial differential equations (PDEs) arise in almost every known physical systems and
at every scale, whether looking at macro scales with continuous mechanics (fluid or solid),
 the very small with quantum physics or
the very large with general relativity. Differential equations constitute also the basis 
of other domains such as finance or population
growth in biology.  In this context, and since the exact solutions of these equations 
are generally not reachable, an efficient and
accurate technique to estimate these solutions is critical. In the last century, the theory and numerical procedures tackling these problems 
have been developed thoroughly.
One of the most successful techniques is called \textit{finite element method} (FEM).
The principle of this method is to discretize the domain in which one needs to solve
the PDE, and to approximate the solution as a linear combination of known, truncated polynomials.
The idea is then to find the best combination
of these functions, a problem which is usually reduced to a linear system of equations, that is tackled efficiently numerically. However, there still are some unresolved issues with these approaches. With large-scale, 
or high dimensional problems, the numerical cost of
classical methods is prohibitive. In industrial applications, with complex geometries and 
various physics interacting with each other, some efforts
need to be made in order to balance model complexity and computational requirements. Hence, 
with the rise of Deep Learning and the broad range of
problems that can be dealt with these techniques efficiently, it seems natural to try and 
develop new Deep Learning solutions for physical applications. In this context, getting real data or accurate simulations is expensive and time-consuming, thus available datasets are rare and scarse. Classical Deep Learning frameworks may thus be innapropriate for this range of problems, therefore, the objective is to develop new techniques combining physical knowledge and the scarse available data to create accurate and trustworthy models.
\\

Many works have focused on this range of problems. With this increasing interest,
some works have been naturally focusing on the dataless, direct solving of partial differential equations with numerical surrogates using Deep Learning
frameworks. Some early works on this topic trace back to the end of the last century, see, for instance
 \cite{lagaris1998artificial}, but the 
subject has really gained attention since a few years ago, with the seminal work of \cite{raissi2019physics}.
 Following this regain in interest, many different techniques and variations of 
the original approach have been developed, often
combining knowledge from numerical analysis and Deep Learning techniques. However, the applications of such techniques
are often limited to 2D cases with regular geometries, settings for which classical numerical methods are more efficient 
and accurate than PINN-based approaches. 

\subsection*{Partial differential equations and PINNs}

For the sake of simplicity, the following section will focus on steady state formalism. A steady PDE can be written in a general form like Equation \eqref{eq:steadyPDE}.

\begin{subequations}
\begin{equation}
    \mathcal{E}(u)(x) = 0 \quad  \forall x \in \Omega, \quad \mathcal{E}\mbox{ being any differential operator,}
\end{equation}

\begin{equation}
    \mathcal{B}(u)(x) = 0 \quad \forall x \in \partial \Omega \quad \mbox{(Boundary conditions, noted (B.C))}.
    \label{eq:BCsteady}
\end{equation}
\label{eq:steadyPDE}
\end{subequations}

\noindent Here, $\mathcal{E}$ represents the differential operator of the equation, it can be linear or nonlinear, 
an elliptic operator for instance. $\Omega$ is the domain in which the equation has to be solved (usually, an open, connex 
set of $\mathbb{R}^d$ for any integer $d \geq 1$). The solution we look for is $u$, function of the variable 
space $x \in \Omega$. Finally, $\mathcal{B}$ is any boundary operator (Dirichlet, Neumann or mixed usually, 
but it can be more general).
\\

The original idea, as developed in \cite{lagaris1998artificial}, updated and improved by \cite{raissi2019physics} to tackle the resolution of steady PDEs such 
as \eqref{eq:steadyPDE} with neural networks was to use a rather simple network architecture, 
the Multi-Layer Perceptron (MLP).
This model produces a predicted solution $\hat{u}$, and in order to ensure that the PDE is respected in the domain by this solution, some \textit{collocation points} $(x_i)_{i = 1,...,N}$ 
are used. These are points chosen, randomly or not, in the domain $\Omega$, for which the 
residuals $\mathcal{E}(\hat{u})(x_i)$ are computed. 
The loss function that needs to be minimized during training is hence:

\begin{equation}
    \mathcal{L}(\hat{u}) = \frac{1}{N} \sum^N_{i=1} ||\mathcal{E}(\hat{u})(x_i)||^2 := \mathcal{L}_r(\hat{u}).
\end{equation}

 A neural network with smooth activation functions is differentiable, so   $\mathcal{E}(\hat{u})$ can be computed, either, for simple cases such as in \cite{lagaris1998artificial}, analytically, or with auto differentiation techniques provided by 
 frameworks such as Pytorch \cite{pytorch2019}. 
 The minimization procedure of the loss is typically done by generalized gradient descent techniques. To ensure that the boundary conditions are respected as well, some collocation 
points are selected on the border as well, to compute the gap between the predicted and expected BC value. The loss is hence: 
\begin{equation}
    \mathcal{L} = \lambda_{r} \mathcal{L}_r(\hat{u}) + \lambda_{u_b}\mathcal{L}_{u_b}(\hat{u}),
    \label{eq:loss-relaxedBC}
\end{equation}

with $\lambda_{r}, \lambda_{u_b}$ some hyperparameters that need to be tuned. Here, supposing there are $M$ data points
located on the border $\partial \Omega$ for which the expected value $u_b$ is known, the loss term $\mathcal{L}_{u_b}$ is:
\begin{equation}
    \mathcal{L}_{u_b}(\hat{u}) = \frac{1}{M} \sum^M_{i=1} ||\hat{u}(x_i) - u_b(x_i)||^2.
    \label{eq:loss-BC}
\end{equation}

Models trained in this framework are known as \textit{Physics-Informed neural networks} (PINNs).

\subsection*{Deep Learning on graphs}

Graph-based approaches have been trending in Deep-Learning, in many fields such as chemistry, 
social interactions modelization or biology. For an overview and some examples of these approaches, see 
\cite{wu2020comprehensive,shivaditya2022graph}. This framework yield state-of-the-art results, which is a consequence of the \textit{inductive biases} of these models. This concept has been studied 
 thoroughly in \cite{battaglia2018relational}. With this structure, the main idea is to propagate information between nodes and edges, step by step, from closer nodes to more distant ones. For this, \textit{graph convolutions}  \cite{kipf2016semi} are used.
\newline

A graph convolution corresponds to a Fourier transform of the graph laplacian matrix, which necessitates the computation of its eigenvalues. However, in order to circumvent this high computational cost, the authors of \cite{kipf2016semi} decided to use a first-order polynomial approximation of this transformation. For the sake of brievty, we will not detail the mathematical formulation of this operation. The underlying idea is to only compute, at each convolutional layer, the influence of the direct neighbours of each node in order to update the predicted value at this node. This operation appears to be very efficient, and to accurately propagate information through the graph. This framework has therefore been widely used in graph-based prediction tasks.
 \newline

 These convolutions are based on the nodes of the graph, but other approaches focus on edge-based information propagation. In \cite{wang2019dynamic}, the authors propose EdgeConv, a convolution based on the edges features of the graph. This architecture has the same inductive biases as nodes convolutions, but also includes the ability to exploit the relative difference of node features. It is particularly suitable for tasks in which the relative position of nodes between themselves, rather than their absolute coordinates, is relevant. It is the case for physical diffusion processes for instance. The authors prove the ability of the model to diffuse local information over the graph, but also to capture more global, long-distance phenomena.

\section{Physics-Informed graph neural networks}

In \cite{chamberlain2021grand}, the message-passing step over a graph is explained as a diffusion process, and the authors prove that the different convolutions can be seen as different techniques to solve a discretized heat diffusion PDE. This strong link between graph convolutional architectures and physical diffusion processes, along with the natural inductive biases explained in \cite{battaglia2018relational}, motivates the use of this framework to simulate physical processes. As a result, several recent works have been focusing on this task.
\newline

In \cite{gao2022physics}, the authors take advantage of the natural link between graphs and meshes to tackle the problem of solving static PDEs. A mesh is seen as a graph, and the neural network learns the discretized solution of the variational problem in each point of the graph. However, only 2 dimensional cases are tackled. \cite{belbute2020combining} incorporated a numerical solver inside a graph model. The result of the solver, which was computed on a coarse mesh, are fed into the graph network to help convergence.

\section{Building PDE residuals on complex geometries}
\label{sec:customGrad}

Many works using the Physics-Informed framework have been focusing on one or two dimensional problems, for which classical numerical techniques already offer fast and accurate simulations. Hence, it seems natural to extend Physics-Informed models to complex three-dimensional geometries, for which classical techniques are slow. However, for these irregular problems, there is an underlying issue with the computation of the PDE residuals, that are necessary to train Physics-Informed models. To prove this issue, let us assume that we have a linear model $M$, which takes as input a field $\varphi$ and the position $x = (x_1, x_2, x_3)$ on a given geometry $\Omega \in \mathbb{R}^3$, and which outputs a field $M(\varphi, x)$. Suppose that the input field $\varphi$, which can be any physical information on the problem, such as the boundary conditions, depends on the position: $\varphi = \varphi(x)$.
\newline
Since $M$ is linear, there exist scalar numbers $\alpha, \beta, \gamma$ such that $M(\varphi, x) = \alpha \varphi(x) + \beta x + \gamma.$ Thus, the corresponding derivative of the output with respect to the variable $x_1$ is:

\begin{equation}
\frac{\partial M}{\partial x_1} = \alpha \frac{\partial \varphi}{\partial x_1} + \beta.
\end{equation}

\begin{figure}[htbp]
    \centering
    \includegraphics[width = 0.8\textwidth]{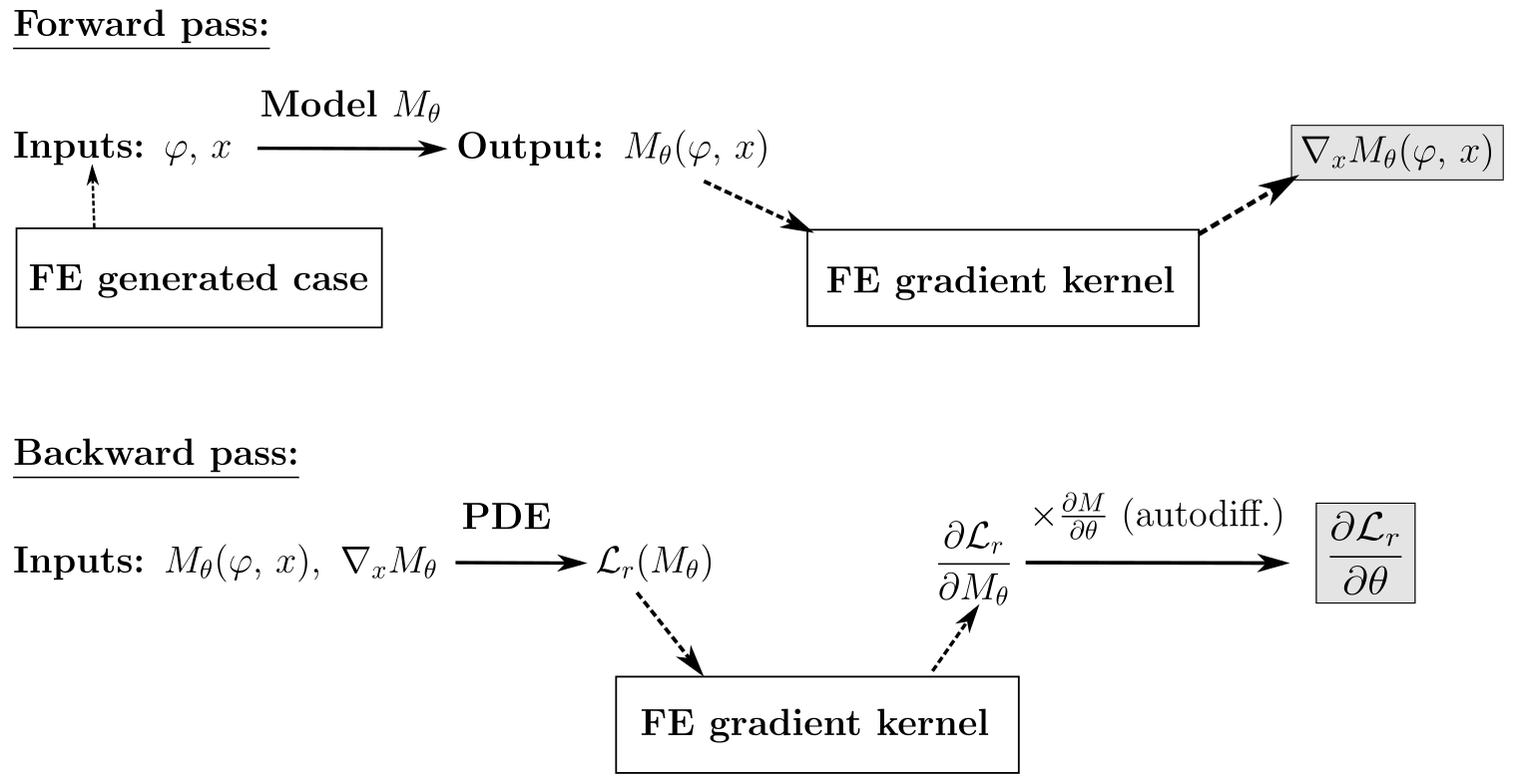}

    \caption{Workflow of the proposed process. The dashed lines represent the exchange of information between the Deep Learning process and the finite element solver, outside of the autodifferentiation framework.}
    \label{fig:workflow}
\end{figure}

For simple, two dimensional cases, or regular three dimensional cases, $\varphi$ can usually be analytically constructed from $x$. But for more complex cases, when dealing with irregular geometries for example, this may not be possible. Hence, the field $\varphi$ may be an input of the model in the form of a spatial field, with no analytical expression. Autodifferentiation \cite{pytorch2019} enables automatic computation of derivatives by applying the chain rule using a recorded computational graph. Every operation on a tensor will be recorded and traced back afterwards, and the actual derivative of this operation will be constructed from these steps, by the chain rule. The derivative of the output with respect to $x_1$ will thus be computed once again as $\frac{\partial M}{\partial x_1} \, \underset{\mbox{autodiff}}{=} \, \alpha \frac{\partial \varphi}{\partial x_1} + \beta.$ However, $\varphi$ being a direct input to the model, there is no recorded link in the computational graph between $\varphi$ and $x_1$: therefore, $\frac{\partial \varphi}{\partial x_1} \underset{\mbox{autodiff}}{=} 0 $. Hence, the derivative of $M$ with respect to $x_1$ will be computed as:
\begin{equation}
    \frac{\partial M}{\partial x_1} \, \underset{\mbox{autodiff}}{=} \, \alpha.
\end{equation}
This issue prevents the use of the Physics-Informed framework for complex geometries, this is why there is a clear lack of work in this direction. There are ways to overcome this issue, for instance, \cite{xiang2022rbf} approximated the spatial derivatives with finite differences, allowing to tackle three dimensional problems. However, this approach does not generalize well for irregular problems, since it is known from classical numerical analysis that finite differences are not accurate enough. Instead, finite element approaches have the flexibility and robustness to allow the computation of approximate spatial derivatives on very irregular classes of geometries. The remaining issue is that these computations are often made inside numerical solvers, and operate complex operations that are not reachable by the Deep Learning autodifferentiation methods. Hence, if an external numerical solver is used to compute the loss residuals, the autodifferentiation will not trace back this computation step, preventing the model to be trained. The missing link is thus the derivative of the numerical gradient operator with respect to the field, i.e the second derivative of the field, the hessian. This second-order derivative could also be computed by a numerical solver. Therefore, in this work, we will be combining the Physics-Informed workflow and a finite element solver to compute the PDE residuals. The corresponding workflow is presented in Figure \ref{fig:workflow}. Note that in this workflow, the model $M_{\theta}$ could be any machine learning model, and is not restricted to a graph network or a neural network.

\section{Proposed approach and results}
\subsection{Problem tackled and chosen model}
\label{subsec:pb}
In order to test the framework presented in the previous section, along with a graph-based approach, an electrostatic physical problem has been chosen. The unknown is the electric potential $V$ on a given, three dimensional, irregular geometry. Two portions of the boundary are fixed, as boundary conditions, and the PDE that needs to be solved for $V$ is simply:

\begin{equation}
    \Delta V = 0.
\end{equation}

\begin{figure}[!ht]
    \centering
    
    \includegraphics[width=0.9\textwidth]{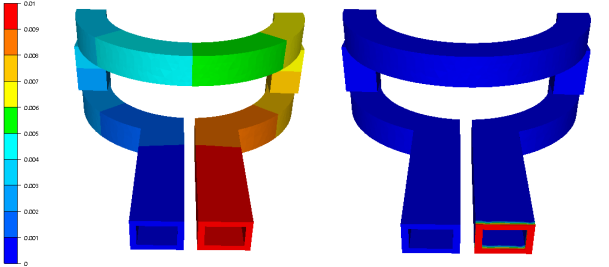}

    \caption{(left) Target field, solution computed by the finite element solver. (right) Boundary conditions field, given as input to the model.}
\label{fig:trueField}
\end{figure}

\noindent This problem has been solved by the commercial finite element solver FORGE \textregistered \cite{alves2017numerical}. In order to combine the classical FE solver with the Physics-Informed workflow two key steps are distinguished: first, mesh and field data created by the software are used as input to contruct the graph-based model. The gradient kernel of the FE solver is used throughout the learning process, as detailed in Section \ref{sec:customGrad}. The target solution, along with the boundary conditions field, can be seen in Figure \ref{fig:trueField}. This graph structure is then fed to the Deep Learning model.

The model is based on the Edge Convolution described in section \ref{sec:intro}. The network is composed of two edge convolution layers, each one being a 2 hidden layers MLP with ReLU activation function, with a width of 128. The model is then trained with the Adam optimizer, with the learning parameters (number of epochs, learning rate, weight decay...) depending on the problem. For the results gathered in table \ref{table:results}, our model is denoted as PIECN, for \textit{Physics-Informed Edge Convolutional Network}.
\newline 

For the loss function, several options have been explored, to validate the results made in the previous sections. The idea being to exploit at its best a single numerical simulation, our approach slightly differs from a pure Physics-Informed approach. The PDE residuals are used for the loss function, along with the mean absolute error (MAE) between the model's prediction and the target field. When two loss terms are used, for the weights between the two terms, several options have been explored. An implementation of \cite{wang2021understanding} has been tried, but there was no significative improvement over fixed weights during training. Hence, for the results presented after, the weights have been fixed to 1 for both terms, except for the PDE residuals computed with the finite element solver, where the weight is set to 0.01. This value is in agreement with the results in the bibliography, since the gradients of this term tend to be larger than for the classical terms.

\subsection{Results and discussions}

In order to test the different aspects of the proposed approach, several models have been tested. Three performance indicators have been monitored: the Mean Absolute Error (MAE), which is the $L_1$ error between the predicted and the actual field, the PDE residuals as computed by the custom spatial gradients, with the procedure described in section \ref{sec:customGrad}, and the PDE residuals as computed by the autodifferentiation module used (here, Pytorch). "PINN" means that the tested model was a MLP, in the same fashion as \cite{raissi2019physics}, and "PIECN" refers to our models, as described in \ref{subsec:pb}. The results are gathered in table \ref{table:results}. 

\begin{table}[htbp]
    \centering
        \begin{tabular}{| c || c | c | c |}
        \hline
        & PDE residuals & MAE &Autodiff. Residuals \\
        \hline
        PINN with autodiff. residuals & 0.111 & 0.950 & 0.0\\
        \hline
        PIECN with autodiff. residuals & 0.078 & 0.769  & $9.38 . 10^{-5}$ \\
        \hline
        PINN trained on MAE & 0.042 &  0.155 & - \\
        \hline
        PIECN trained on MAE & $2.66 . 10^{-4}$ & 0.0082 & - \\
        \hline
        PINN, spatial residuals + MAE & 0.034 & 0.180 & - \\
        \hline
        PIECN, spatial residuals + MAE & $\mathbf{2.19 . 10^{-4}}$ &  $\mathbf{0.0070}$ &  - \\
        \hline
        \hline
        Target solution & 0.0 & 0.0 & - \\
        \hline
    \end{tabular}
    
    \caption{Results of the tested models on the electrostatic problem.}
    \label{table:results}
\end{table}

First, with the two first lines, it is clear that the residuals computed by autodifferentiation are not suitable. With this framework, a PINN achieves a zero residual, without actually converging to the true solution, and same goes for the PIECN. On the other hand, as seen in the last line, the target solution has a zero residual when computed with our own gradient implementation. As explained in section \ref{sec:customGrad}, our approach circumvents the issue of autodifferentiation. 
\newline

Overall, it is clear that a simple neural network is less efficient than graph-based approaches. The inputs of the PINN are the absolute positions and fields, although for such physical processes, the relative positions between points are more significant. The diffusion process happens inside the geometry considered, and is well captured by the mesh-based geometry of our approach, while the PINN is not able to capture this phenomenon. Finally, for the graph model, a training on the field and the residuals of the equations yields slightly more accurate results than a training on the field only. The difference is not spectacular, but this may be due to the smoothness of the equation considered, and the fact that the geometry is not very complex. Some further work on more complex physical problems will be conducted in order to quantify the benefit of this approach.
\newline

\begin{figure}[htbp]
    \centering
    \includegraphics[scale = 0.6]{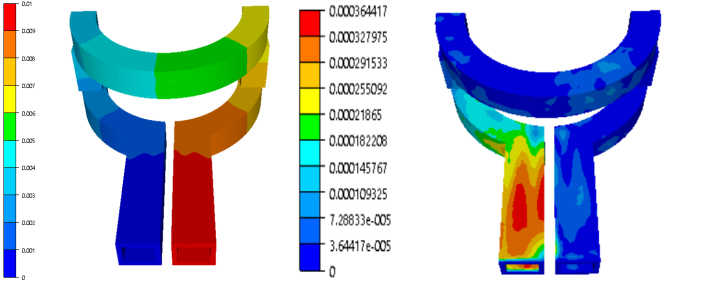}
    \caption{(left) Prediction of our model. (right) Corresponding MAE.}
    \label{fig:predField}
\end{figure}

The prediction made by the most accurate model, the PIECN trained on both the field and the PDE residuals, has been reproduced in Figure \ref{fig:predField}, along with the corresponding MAE. The peak errors are located farthest from the non-zero boundary condition, in agreement with the way diffusion happens on a graph convolutional model.

\vspace{-0.1cm}

\section{Concluding remarks}

In this work, after an overview of the works tackling Physics-Informed Deep Learning for physical problems, we provided insight on why graph-based learning are well suited for these applications. We then proved a limitation of autodifferentiation techniques for Physics-Informed learning, and we proposed a new workflow combining a finite element solver and Deep Learning to tackle this issue. Finally, we proved the benefit of this approach on a three dimensional geometry. This new framework will allow us to tackle more challenging cases in the future. First, a quantification of the benefit of combining physial knowledge and numerical data should be conducted. Then, the generalization capability of such models should be investigated. Finally, the proposed approach should be tested on more complex equations such as non linear, time-dependent problems.

\bibliography{bibliography}
\bibliographystyle{abbrv}

\end{document}